\documentclass[10pt,twocolumn,letterpaper]{article}

\usepackage[pagenumbers]{iccv}      
\usepackage{multirow}
\usepackage{siunitx}
\usepackage{makecell}
\usepackage{mathptmx}
\definecolor{iccvblue}{rgb}{0.21,0.49,0.74}
\usepackage[pagebackref,breaklinks,colorlinks,allcolors=iccvblue]{hyperref}

\def\paperID{5127} 
\def\confName{ICCV}
\def\confYear{2025}

\title{UniGlyph: Unified Segmentation-Conditioned Diffusion for Precise Visual Text Synthesis}

\author{
Yuanrui Wang$^{1,2,\ast}$, 
Cong Han$^{2,\ast}$, 
Yafei Li$^{2}$, 
Zhipeng Jin$^{2}$, 
Xiawei Li$^{2}$, 
SiNan Du$^{1}$,\\
Wen Tao$^{2}$, 
Shuanglong Li$^{2}$, 
Yi Yang$^{2,\dagger}$, 
Chun Yuan$^{1,\dagger}$, 
Liu Lin$^{2}$ \\
$^{1}$Tsinghua University \qquad $^{2}$Baidu Inc. \\
{\tt\small wang-yr23@mails.tsinghua.edu.cn, hancong01@baidu.com, liyafei03@baidu.com} \\
{\tt\small dsn23@mails.tsinghua.edu.cn, jinzhipeng@baidu.com, lixiawei01@baidu.com, taowen02@baidu.com} \\
{\tt\small lishuanglong@baidu.com, yangyi15@baidu.com, liulin03@baidu.com}
}

\begin{document}
\maketitle
\begin{abstract}
Text-to-image generation has greatly advanced content creation, yet accurately rendering visual text remains a key challenge due to blurred glyphs, semantic drift, and limited style control. Existing methods often rely on pre-rendered glyph images as conditions, but these struggle to retain original font styles and color cues, necessitating complex multi-branch designs that increase model overhead and reduce flexibility.
To address these issues, we propose a segmentation-guided framework that uses pixel-level visual text masks—rich in glyph shape, color, and spatial detail—as unified conditional inputs. Our method introduces two core components: (1) a fine-tuned bilingual segmentation model for precise text mask extraction, and (2) a streamlined diffusion model augmented with adaptive glyph conditioning and a region-specific loss to preserve textual fidelity in both content and style.
Our approach achieves state-of-the-art performance on the AnyText-benchmark, significantly surpassing prior methods in both Chinese and English settings. To enable more rigorous evaluation, we also introduce two new benchmarks: GlyphMM-benchmark for testing layout and glyph consistency in complex typesetting, and MiniText-benchmark for assessing generation quality in small-scale text regions. Experimental results show that our model outperforms existing methods by a large margin in both scenarios, particularly excelling at small text rendering and complex layout preservation, validating its strong generalization and deployment readiness.
\end{abstract}

\renewcommand{\thefootnote}{} 
\footnotetext{$\ast$ Equal Contribution.}
\footnotetext{$\dagger$ Corresponding Authors.}
\section{Introduction}
\label{sec:intro}

In recent years, deep learning has achieved impressive success in various domains of machine learning and artificial intelligence, including but not limited to computer vision, image generation, image processing, and controllable synthesis~\cite{DALLE3,Patrick_SD3_ICML24,FLUX.1,wei2025perceiveunderstandrestorerealworld,zhuang2023task,zhuang2024colorflow,ma2025followcreation,ma2025followyourmotion,ma2024followyouremoji,ma2025followyourclick,wei2025modeling,cheng2025whoever,guan2025enhancinglogitsdistillationplugplay,xu2023chartbench,xu2025chartmoe,sun2024gsrenderdeduplicatedoccupancyprediction,du2024alore}. However, generating precise visual text (glyphs) remains an unsolved challenge, with persistent issues including blurred character edges, semantic inconsistencies, and poor control over typographic attributes like font styles and colors. Current solutions addressing this problem predominantly adopt three paradigms: (1) inpainting-based methods (e.g., Text-Diffuser~\cite{Chen_TextDiffuser_Corr23}), which struggle with text-background harmonization; (2) embedding-injection approaches (e.g., Glyph-ByT5-v2~\cite{glyphbyt5-v2}), limited by predefined typographic constraints; and (3) ControlNet-based architectures, exemplified by state-of-the-art methods AnyText~\cite{tuo2023anytext} and GlyphDraw2~\cite{glyphdraw2}, which currently set the performance benchmark but suffer from fundamental limitations.

\noindent The ControlNet-based paradigm, while effective, relies on pre-rendered glyph images that inherently discard critical typographic information. Specifically, these rendered conditions preserve only glyph shapes and positions while losing original font characteristics and color details. This incomplete signal forces models to learn implicit mappings between synthetic glyphs (typically in default fonts) and real-world typographic variations during training. To compensate, existing methods introduce auxiliary modules for positional encoding (e.g., AnyText's text embedding replacement) and font/color control (e.g., GlyphDraw2's style guidance), resulting in multi-branch architectures that increase computational complexity, reduce model reusability, and create optimization conflicts—particularly when generating small or stylized glyphs. We identify this information degradation in conditioning signals as the root cause of current methods' suboptimal performance and architectural bloat.

\noindent To address these limitations, we propose UniGlyph, a streamlined framework that replaces rendered glyph images with pixel-accurate visual text segmentation masks as unified conditioning signals. Our key insight is that segmentation masks inherently preserve all glyph attributes—shape, position, font style, and color—at full spatial resolution, eliminating the need for auxiliary control modules. The framework comprises three innovations: (1) A bilingual text segmentation module producing typography-preserving masks from source images; (2) A diffusion model enhanced with adaptive glyph condition and glyph region loss to ensure accuracy of glyph generation; (3) An optional layout generation module enabling automatic text arrangement without compromising architectural simplicity. Using a segmentation-guided visual text generation framework, our single ControlNet architecture achieves state-of-the-art performance.

\noindent Training such a system requires high-quality bilingual data that existing datasets fail to provide. We contribute GlyphMM-3M—a Chinese-English text-image dataset (3M+ images) with pixel-aligned glyph annotations—and Poster-100K, a curated dataset for aesthetic typographic generation. Two evaluation benchmarks are introduced: GlyphMM-Benchmark for holistic quality assessment and MiniText-benchmark focusing on small/complex glyphs. Experiments across these benchmarks demonstrate that UniGlyph outperforms prior arts while maintaining superior image quality, validating our hypothesis that complete glyph signal preservation enables both architectural simplification and performance gains.

\begin{figure*}[t]
    \centering
    \includegraphics[width=1\linewidth]{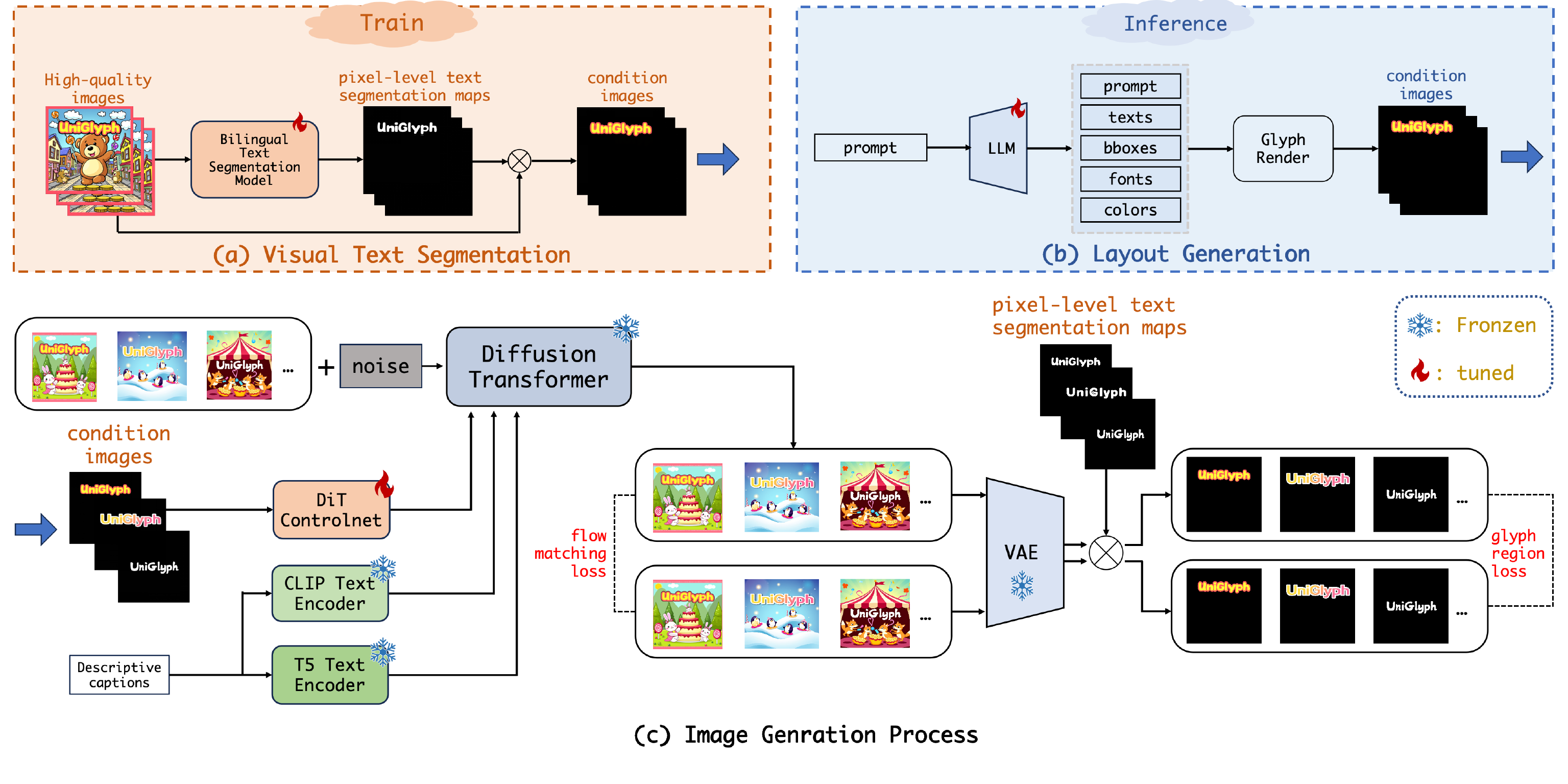}
    \caption{The UniGlyph framework: (1) A dual-objective diffusion model (flow matching + glyph region loss) ensures glyph feature alignment; (2) A visual text segmentation module extracts pixel-level text segmentation masks during training; (3) An optional layout generator enables adaptive text arrangement during inference. This design maintains architectural simplicity while supporting flexible operational configurations.}
    \label{fig:pipeline}
\end{figure*}
\section{Related Work}
\label{sec:relatedwork}

\textbf{Controllable Text-to-Image Diffusion Models}
Text-to-image (T2I) diffusion models have advanced significantly but often lack precise control when relying solely on textual input. ControlNet\cite{Zhang_ControlNet_Corr23}, T2I-Adapter\cite{Mou_T2Iadapter_Corr23}, Composer\cite{Huang_Composer_Corr23}, IP-Adapter\cite{ip-adapter}, and ReferenceNet\cite{hu_animateanyone} are effective methods for incorporating additional control signals into text-to-image diffusion models. However, when it comes to fine-grained control in image generation, ControlNet stands out as the most effective and widely adopted approach.
ControlNet\cite{Zhang_ControlNet_Corr23} enhances the controllability of T2I models by adding an auxiliary encoder connected via zero convolution. This design allows the model to incorporate additional conditioning inputs like depth maps, pose images, and sketches, improving control over the generated images without overfitting.
Due to its effectiveness and adaptability, ControlNet has become foundational in enhancing control within T2I models, influencing numerous subsequent works. For instance, GlyphControl\cite{Yang_GlyphControl_Corr23} first leverages ControlNet for precise visual text rendering by conditioning on glyph shapes, and Control3D\cite{chen2023control3d} uses it to guide diffusion models with 3D information for accurate spatial representations.

\noindent \textbf{Visual Text Generation}
Generating images with accurate and coherent text is a significant focus in generative models, heavily influenced by the text encoder’s capability to render text faithfully. Liu et al.\cite{Liu_CharacterAware_ACL23} emphasized the importance of character-level encoders for capturing word spelling and character appearance. Building on this, various methods have been proposed, categorized as follows:
Methods like DiffSTE\cite{diffste}, UDiffText\cite{zhao_udifftex}, Glyph-ByT5\cite{glyphbyt5}, and SceneTextGen\cite{zhangli2024layoutagnosticscenetext} employ character-level encoders to enhance text representation and rendering fidelity within images.
Approaches such as AnyText\cite{tuo2023anytext,tuo2024anytext2}, GlyphDraw2\cite{glyphdraw2}, TextDiffuser\cite{Chen_TextDiffuser_Corr23}, and GlyphControl\cite{Yang_GlyphControl_Corr23} integrate visual features like pre-rendered glyph images, stroke information, and glyph layout masks to guide accurate text generation.
To improve spatial accuracy and address text placement issues, methods such as Brush Your Text\cite{zhang2023_brushyourtext} and UDiffText impose constraints on attention maps within glyph regions, enhancing positional coherence.
Techniques such as TextDiffuser-2\cite{Chen_TextDiffuser2}, ARTIST\cite{zhang2024artist}, CustomText\cite{paliwal2024customtext}, and Glyph-ByT5\cite{glyphbyt5} offer control over attributes like text layout, font style, and color, enabling personalized and sophisticated text rendering.

\noindent \textbf{LLM for layout generation}
Recent research has revealed that large language models possess capabilities extending beyond traditional language tasks, including layout planning for image generation \cite{touvron2023llama,touvron2023llama2}. Some works, such as LayoutGPT \cite{feng2023layoutgpt} and LayoutPrompter \cite{lin2023layoutprompter}, utilize LLMs to generate layouts in front-end languages like HTML. Other works, such as TextDiffuser-2\cite{Chen_TextDiffuser_Corr23} and GlyphDraw2\cite{glyphdraw2}, use LLMs to generate bounding boxes for rendering text. This inspired us to explore using LLMs to generate other elements besides bounding boxes, such as fonts and colors.
There are two primary approaches to leveraging LLMs for layout generation: prompt engineering with advanced proprietary models or fine-tuning open-source LLMs. Following GlyphDraw2, we chose to fine-tune an open-source LLM; we constructed our own dataset for fine-tuning and found that the fine-tuned LLM adapts well to our task.

\section{Methodology}
\label{sec:method}

The overall framework of UniGlyph comprises several key components, as in Fig.~\ref{fig:pipeline}: (1) Bilingual Text Segmentation Model fine-tuned from\cite{hi-sam_tpami24}: Used during training to obtain text segmentation maps from given images. (2) Flow-Matching based Diffusion Transformer Model and Related Components: This includes a pre-trained VAE, a text encoder and Diffusion Transformer. (3) DiT ControlNet: Utilized to encode visual text signals extracted from the original images via the text segmentation maps. (4) LayoutTransformer and Glyph Renderer: During inference, the LayoutTransformer derives text layout and style information from user prompts. The glyph renderer uses this information to render a text image, which serves as a condition for generation.
\subsection{Preliminary}
\begin{figure}[h]
    \vspace{-0.4cm}
    \centering
    \includegraphics[width=1\linewidth]{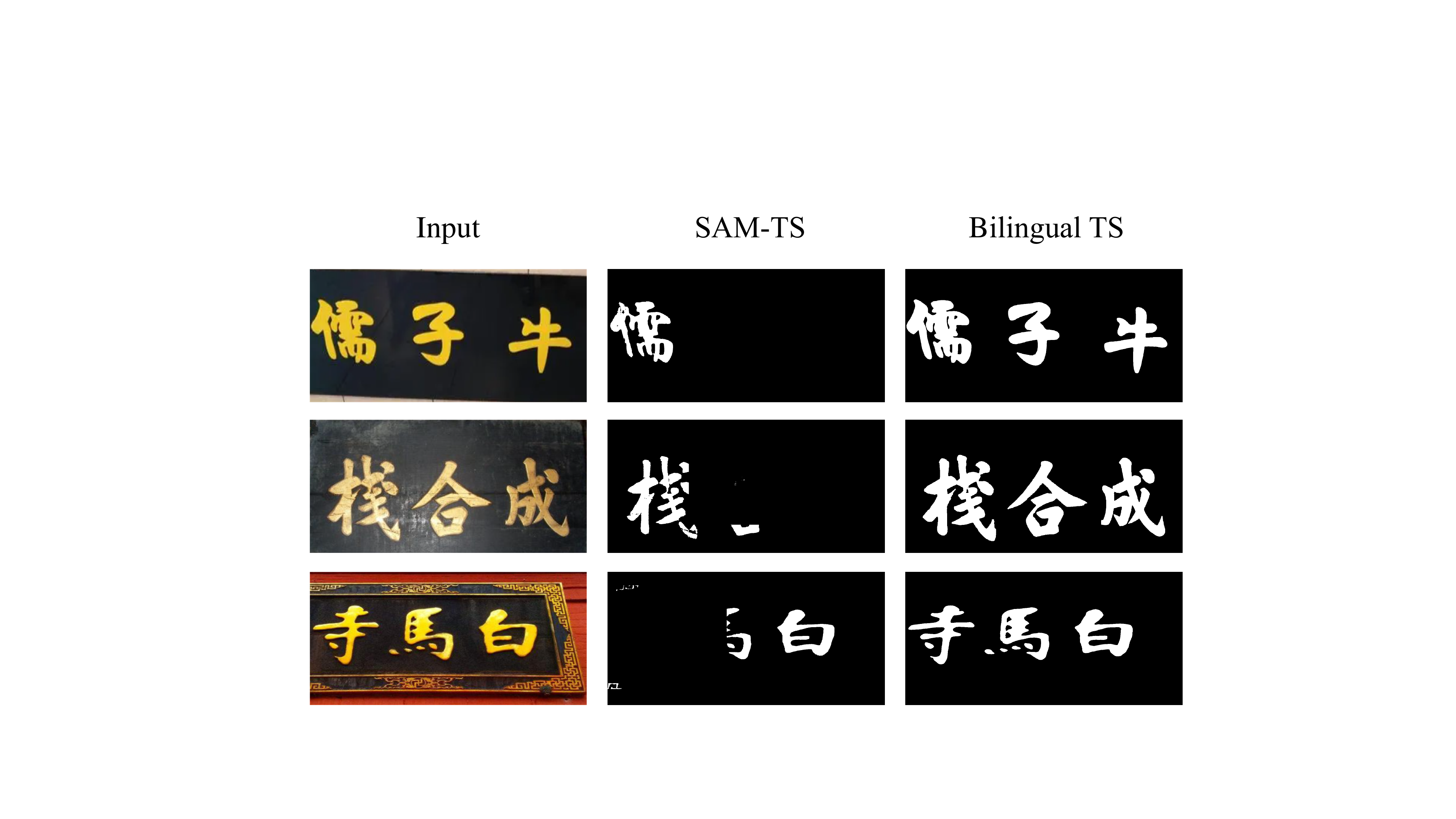}
    \caption{The segmentation results are presented in three columns: the original input images (left), outputs from SAM-TS \cite{hi-sam_tpami24} (middle), and results from our fine-tuned text segmentation model (right).}
    \label{fig:sam-ft}
\end{figure}

{\bf Hi-SAM}\cite{hi-sam_tpami24} is a unified model leveraging SAM\cite{kirillov2023segany} for hierarchical text segmentation, consisting of two parts. The first part, SAM-TS, includes an image encoder from SAM, a plug-in self-prompting module, and a pixel-level text mask decoder, performing pixel-level text segmentation.
The second part functions to predict word, text-line, and paragraph-level segmentation masks for layout analysis based on the first part’s results.
Since Hi-SAM was trained only on English datasets, it originally segments only English text. Our goal is to achieve accurate pixel-level text segmentation for both English and Chinese. Therefore, we fine-tuned only the first part, SAM-TS, using the bilingual (Chinese and English) segmentation dataset BTS\cite{xu2022bts}, resulting in a bilingual Text Segmentation model, as in Fig .\ref{fig:sam-ft}, which we utilize to get the pixel-level segmentation mask:
\begin{equation}
    \textrm{M}_\textrm{seg} = \textrm{SAM-TS}(\textit{I})
\end{equation}
\subsection{Adaptive Glyph Condition}
\begin{figure}[h]
    \centering
    \includegraphics[width=1\linewidth]{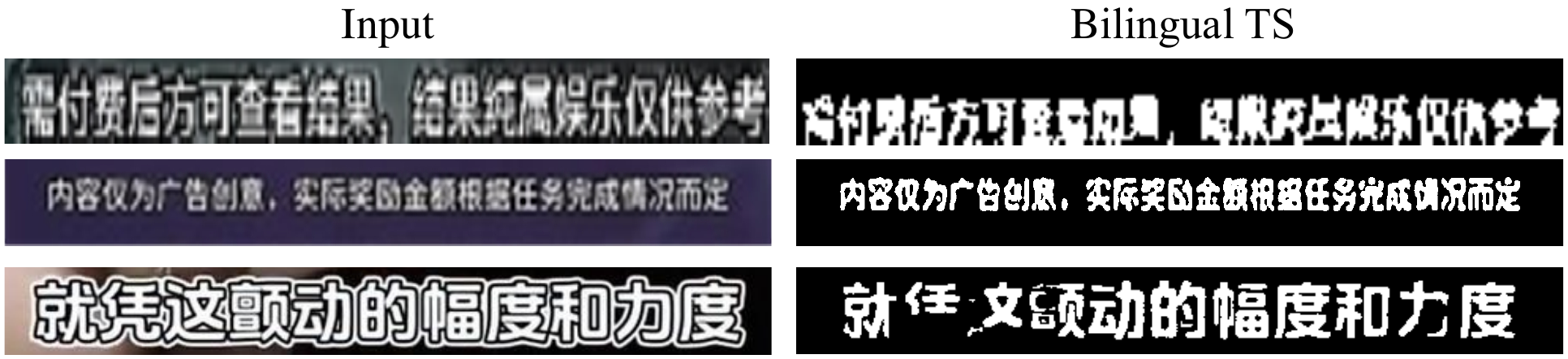}
    \caption{The segmentation performance on extremely compact character instances, as processed by our bilingual text segmentation model, exhibits inaccuracies in segmenting small characters.}
    \label{fig:smallglyph}
\end{figure}

Upon obtaining the segmentation mask, we can derive glyph information from the input image. which contains comprehensive glyph information including shape, position, font, and color. Pixel values are zero in regions outside the glyph, which means if the glyph in $\textit{I}$ is black, it will blend into the background. To address this, we extract the edges from the segmentation map $\textrm{M}_\textrm{seg}$ using the Canny edge detector, we then enhance glyph information by adding the edges to distinguish the glyph from the background. Moreover, our findings indicate that the fine-tuned SAM-TS\cite{hi-sam_tpami24} exhibits inaccuracies in segmenting small characters, as shown in Fig.~\ref{fig:smallglyph}. To be specific, we can only get precise segmentation mask when the glyph region is relatively large. So we implement an adaptive blending strategy based on the size of the glyph regions. Using PP-OCRv4\cite{PP-OCRv4}, we obtain the bounding boxes of glyph regions. Thus we can obtain a position mask $\textrm{M}_\textrm{pos}$ of each glyph region $R_i$ set to 1 with others 0 and calculate areas of $R_i$. For each $R_i$ having an area $A_i$ and containing $N_i$ characters, we compute the average area occupied by a single character:
\begin{equation}
    A_\textrm{avg,i}=\frac{A_i}{N_i}
\end{equation}
We set a threshold $T=4900$ pixels. So the glyph information $\textrm{G}_i$ for each glyph region $R_i$ is then defined as:
\begin{equation}
\textrm{G}_i = \begin{cases}
\textit{Canny}(\textrm{M}_\textrm{seg})+\textrm{M}_\textrm{seg} \odot \textit{I}, & \text{if } A_{\text{avg}, i} > T \\
(\textrm{M}_\textrm{pos}\odot\textit{I}_i)^{\text{blur}}, & \text{if } A_{\text{avg}, i} \leq T
\end{cases}
\end{equation}
\noindent Where $\textit{I}_\textrm{pos}^\textrm{blur} = (\textrm{M}_\textrm{pos}\odot\textit{I}_i)^{\text{blur}}$ is the corresponding cropped region from the original image $\textit{I}$ with Gaussian blur applied to its boundary to smooth the transition between text and non-text areas. The Gaussian blur applied to the boundary of small glyph boxes is defined as $\textit{I}_\textrm{pos}^{\text{blur}} = \textit{I}_\textrm{pos} * \textit{Gaussain}(\sigma)$, where $*$ denotes convolution.\\
\noindent The final condition input $\textrm{G}$ is constructed by combining all $\textrm{G}_i$:
\begin{equation}
    \textrm{G}=\bigcup_i \textrm{G}_i
\end{equation}

\noindent This adaptive approach mitigates signal degradation in glyph regions and ensures seamless integration of the glyph with the generated background.
\subsection{Flow-Based Diffusion Model with condition}
In the flow-based diffusion model\cite{lipman2023flowmatching, wang2024rectifieddiffusion, Patrick_SD3_ICML24} with condition, both input image and glyph condition image are encoded from pixel space $ \textit{I},\textrm{G} \in \mathbb{R}^{H \times W \times 3} $ into latent representation $z_0, z_g \in \mathbb{R}^{h \times w \times c}$ by VAE\cite{Diederik_VAE_ICLR14}. Subsequently, latent diffusion algorithms progressively add noise to input latent $z_0$ over time steps $t \in [0,T]$, resulting in a sequence of noisy latent variables ${z_t}$. In our flow-based diffusion model, we aim to learn a continuous-time velocity field $\mathbf{v}^*(\textbf{\textit{z}}_t, t)$, that transports the data distribution to a prior distribution, effectively modeling the dynamics of the diffusion process. Given latent, including the time step $t$, the glyph feature $z_s = C(z_g,c_{te},t)$ produced by ControlNet, and the text embedding $c_{te}$ from text encoder, the flow-based diffusion model employs the neural network $\textbf{v}_\theta$  to predict the velocity field at each time step. The objective is to match the model’s velocity field to the ideal velocity field that transports the data distribution along the diffusion process. This is achieved by minimizing the flow matching loss:
\begin{equation}
\textrm{L}_{\textrm{fm}} = \mathbb{E}_{\textbf{\textit{z}}_0, \textbf{\textit{z}}_s, \textbf{\textit{c}}_\textrm{te}, t} \left[ \left|\left| \mathbf{v}_{\theta}(\textbf{\textit{z}}_t, \textbf{\textit{z}}_s, \textbf{\textit{c}}_\textrm{te}, t) - \mathbf{v}^*(\textbf{\textit{z}}_t, t) \right|\right|_2^2 \right]
\end{equation}
\noindent where $L_{\textrm{fm}}$ is flow matching loss, and $ \mathbf{v}^*(\textbf{\textit{z}}_t, t) $ is the target velocity field derived from the diffusion process that ideally transports $z_t$ over time. 

\subsection{Glyph Region Loss}
To further enhance the accuracy of text generation in images, we propose a glyph region loss that leverages $\textrm{M}_\textrm{seg}$ to refine the synthesis of glyph content. During training, we reconstruct the original image $\hat{\textbf{x}}_0$ in pixel space by utilizing the predicted velocity field $\textbf{v}_\theta$ and the VAE decoder. We then compute the Mean Squared Error (MSE) loss between these regions to specifically focus on the fidelity of the glyph generation. This is equivalent to assigning an additional weight to the glyph region, which enhances the model’s focus on the glyph itself. Following the settings from the previous section, we obtain our mask map as follows:
\begin{equation}
\textrm{M}_\textrm{gr} = \begin{cases}
\textrm{M}_\textrm{seg}, & \text{if } A_{\text{avg}, i} > T \\
\textrm{M}_\textrm{pos}, & \text{if } A_{\text{avg}, i} \leq T
\end{cases}
\end{equation}
As a result we get our glyph region loss as:
\begin{equation}
\textrm{L}_{\textrm{gr}} = \mathbb{E}_{\mathbf{x}_0, \hat{\mathbf{x}}_0} \left[ \left|\left| \textrm{M}_\textrm{gr} \odot (\hat{\mathbf{x}}_0 - \mathbf{x}_0) \right|\right|_2^2 \right]
\end{equation}
We assigned a weight $\lambda$ to the glyph region loss and discussed its value in Table~\ref{tab:alation-lambda}. Finally, we obtain the loss for our task.
\begin{equation}
    \textrm{L} = \textrm{L}_\textrm{fm} + \lambda * \textrm{L}_\textrm{gr}
\end{equation}

\subsection{LLM for layout prediction}
We fine-tuned a large language model to convert user prompts into layout and style information. Specifically, the input to the task is the user’s prompt \texttt{<prompt>}, and the output is a structured format comprising \texttt{<rewritten prompt, texts, bboxes, fonts, colors>}, where the rewritten prompt no longer includes the character information to be rendered. To constrain the fonts and colors generated by the model, we predefined sets of fonts and colors, mapping them to special tokens formatted as \texttt{<font\_xxx>} and \texttt{<color\_xxx>}, which were added to the tokenizer’s vocabulary. We constructed 1,000 pieces of structured data using poster image data for fine-tuning. After fine-tuning, we found that the model could understand the semantic relationships among the structured data. For example, the model does not generate overlapping bboxes for different two-line texts, and the sizes of the bboxes generated by the model are often proportional to the number of characters contained in the text.
\vspace{-0.2cm}
\section{Dataset and Benchmark}
\label{sec:datasetandbenchmark}
\subsection{Dataset}
\textbf{Anyword}: A dataset proposed in AnyText\cite{tuo2023anytext}, comprising millions of Chinese and English text images from Wukong\cite{wukong_corr22} (Chinese) and LAION\cite{laion_400m_corr21}(English). All images have a resolution of $512\times512$ pixels. However, the background quality of the Chinese glyph images is generally poor.

\noindent \textbf{GlyphMM-3M}: To achieve high-quality visual text generation, we constructed a large-scale, high-resolution bilingual text image dataset. This dataset contains over 3 million text images, including both natural scene images and poster images. It includes images of various resolutions, enabling the model to learn text layout and distribution at different resolutions during training, and to adapt to text image generation at various resolutions during inference. The dataset provides segmentation maps, captions, and OCR annotations, allowing for model training in various scenarios. 

\noindent \textbf{Poster-100K}: This dataset contains more than 100,000 high-quality poster images, most of which are in Chinese with a small portion in English. All images are in portrait orientation. The posters predominantly feature multi-line glyph, and the background images are more complex and of higher quality, aiming to enhance the model’s ability to render aesthetically pleasing posters.
\vspace{-0.2cm}
\subsection{Benchmark}
\textbf{AnyWord-Benchmark}\cite{tuo2023anytext}: This dataset contains 1,000 English images and 1,000 Chinese images sourced from LAION and Wukong, respectively.

\noindent \textbf{GlyphMM-benchmark}: This dataset comprises 1,200 Chinese images and 1,000 English images. It includes a large number of text images featuring complex glyphs and multi-line text. OCR annotations are also provided.

\noindent \textbf{MiniText-benchmark}: This dataset consists of 400 manually annotated Chinese images containing small-sized text, used to evaluate the model’s ability to generate small Chinese glyph.

\begin{figure*}[t]
    \centering
    \includegraphics[width=1\linewidth]{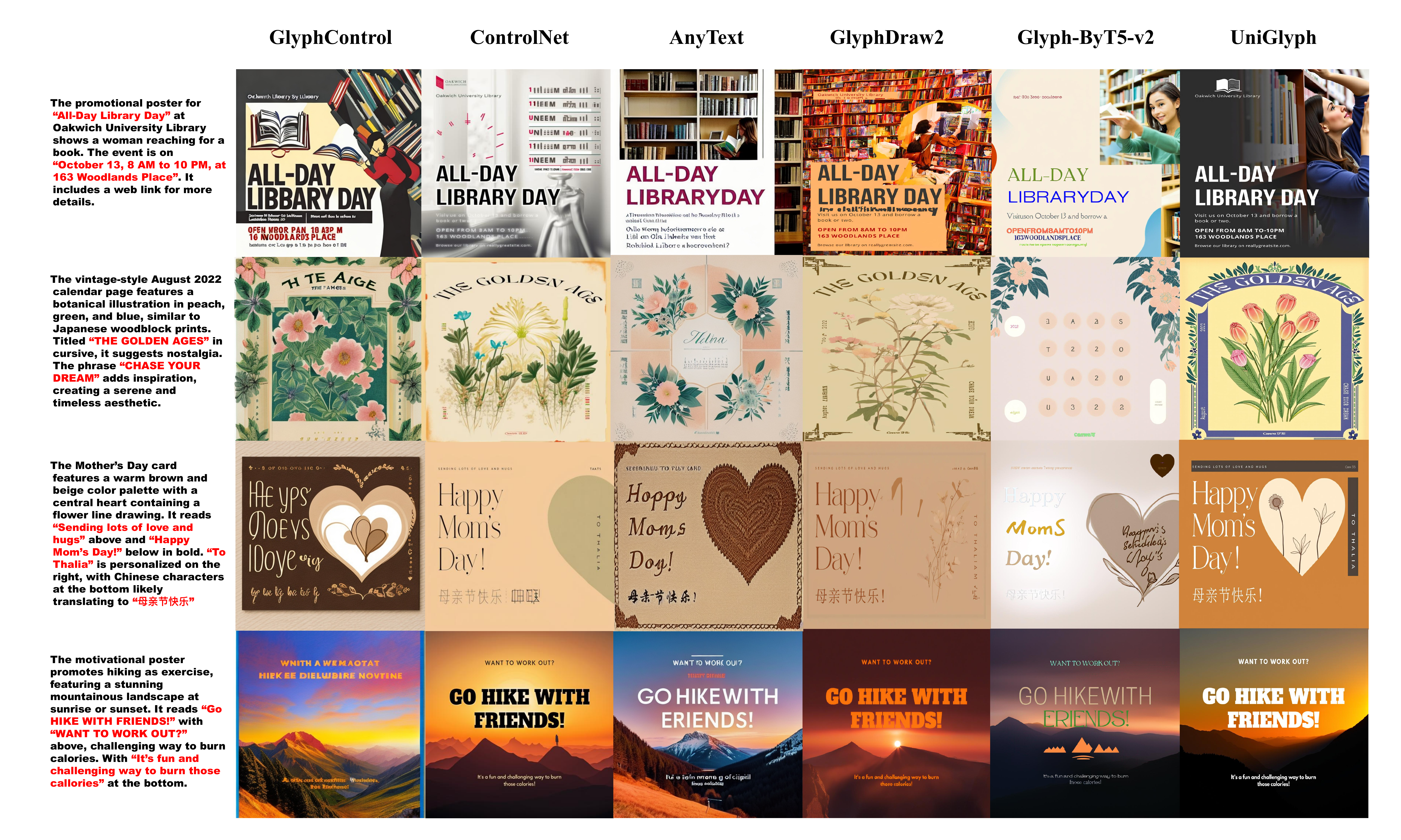}
    \caption{Qualitative comparison between UniGlyph and state-of-the-art models, primarily illustrating English text generation, with the third row showcasing Chinese text generation. All captions are selected from the English evaluation in the GlyphMM-benchmark.}
    \label{fig:qualires}
\end{figure*}
\section{Experiment}
\label{sec:experiment}
\begin{table*}[t]
\centering
\caption{Experimental results on AnyText-benchmark and GlyphMM-benchmark.}
\sisetup{
  output-decimal-marker={.},
  table-format=1.4,
}
\begin{tabular}{ll *{4}{S} *{4}{S}}
\toprule
  \multicolumn{2}{c}{} & \multicolumn{4}{c}{Chinese} & \multicolumn{4}{c}{English} \\
  \cline{3-6} \cline{7-10}
  \multicolumn{1}{c}{\multirow{-2}{*}{Benchmark}} & \multicolumn{1}{c}{\multirow{-2}{*}{Method}} & {Sen.Acc} & {NED} & {ClipScore} & {HPSv2} & {Sen.Acc} & {NED} & {ClipScore} & {HPSv2} \\
  \hline
  \multirow{8}{*}{\makecell{AnyText-\\benchmark}} 
& SD3 & 0.0020 & 0.0074 & 0.7681 & 0.2740 & 0.3261 & 0.4422 & 0.7321 & 0.2215 \\
& ControlNet & 0.3620 & 0.6227 & 0.8335 & 0.2347 & 0.5837 & 0.8015 & 0.7711 & 0.2245 \\
& GlyphControl & 0.0454 & 0.1017 & 0.7974 & 0.2579 & 0.5262 & 0.7529 & 0.8160 & 0.2566 \\
& AnyText-V1.1 & 0.6923 & 0.8423 & 0.8341 & 0.2272 & 0.6564 & 0.8685 & 0.8822 & 0.2121 \\
& GlyphDraw2 & 0.7350 & 0.8451 & 0.8204 & 0.2589 & 0.7369 & 0.8921 & 0.7974 & 0.2451 \\
& Glyph-ByT5 & 0.6938 & 0.8197 & 0.7905 & 0.2601 & 0.6225 & 0.7658 & 0.8345 & 0.2511 \\
& AnyText2 & 0.7130 & 0.8516 & 0.8139 & \multicolumn{1}{c}{--} & 0.8096 & 0.9184 & 0.8963 & \multicolumn{1}{c}{--} \\
& CharGen\cite{ma2024chargenhighaccuratecharacterlevel} & 0.7499 & 0.8609 & \multicolumn{1}{c}{--} & \multicolumn{1}{c}{--} & 0.8096 & 0.9205 & \multicolumn{1}{c}{--} & \multicolumn{1}{c}{--} \\
& UniGlyph & \textbf{0.8267} & \textbf{0.8976} & 0.7915 & 0.2551 & \textbf{0.9018} & \textbf{0.9582} & 0.8655 & 0.2473 \\
  \hline
  \multirow{7}{*}{\makecell{GlyphMM-\\benchmark}}
& SD3 & 0.0001 & 0.0021 & 0.7867 & 0.2052 & 0.0000 & 0.0224 & 0.8271 & 0.2012 \\
& ControlNet & 0.0082 & 0.0331 & 0.7983 & 0.2407 & 0.0001 & 0.0134 & 0.8479 & 0.2273 \\
& GlyphControl & 0.0007 & 0.0112 & 0.6640 & 0.2193 & 0.0122 & 0.2231 & 0.9024 & 0.1969 \\
& AnyText-V1.1 & 0.3004 & 0.4656 & 0.6429 & 0.2442 & 0.1718 & 0.4998 & 0.8556 & 0.2095 \\
& GlyphDraw2 & 0.4232 & 0.5779 & 0.6159 & 0.2456 & 0.3079 & 0.6564 & 0.8512 & 0.1958 \\
& Glyph-ByT5 & 0.4300 & 0.6150 & 0.8413 & 0.2178 & 0.2812 & 0.5952 & 0.8673 & 0.2154 \\
& UniGlyph & \textbf{0.5056} & \textbf{0.6577} & 0.8216 & 0.2522 & \textbf{0.4522} & \textbf{0.8179} & 0.8768 & 0.1987 \\
\bottomrule
\end{tabular}
\label{tab:resultsontwobenchmark}
\end{table*}

\begin{table}[htbp]
\centering
\caption{Experimental results on MiniText-Benchmark.}
\label{tab:results}
\begin{tabular}{@{}lccc@{}}
\toprule
Method & Sen.Acc & NED & ClipScore \\
\midrule
SD3 & 0.0000 & 0.0005 & 0.7990 \\
ControlNet & 0.0006 & 0.0021 & 0.7958 \\
GlyphControl & 0.0000 & 0.0046 & 0.8075 \\
AnyText-V1.1 & 0.0138 & 0.4680 & 0.8098 \\
GlyphDraw2 & 0.0100 & 0.4508 & 0.8146 \\
Glyph-ByT5 & 0.3881 & 0.8268 & 0.8594 \\
\midrule
UniGlyph & \textbf{0.7925} & \textbf{0.9537} & 0.8124 \\
\bottomrule
\end{tabular}
\label{tab:resultsonminitext}
\end{table}

\subsection{Implementation Details}
\subsubsection{Fine-tuning of Segmentation Model}
Following Hi-SAM\cite{hi-sam_tpami24}'s default training settings, we fine-tuned the SAM-TS model on the BTS dataset\cite{xu2022bts}. Our training focused on three key components: the image encoder adapter, self-prompting module, and S-Decoder, comprising a total of 57.96 million trainable parameters. The complete system was trained for 70 epochs under this configuration.
\subsubsection{Training of Diffusion Model}
Our diffusion model training framework is based on DiT\cite{peebles2023dit} and DiT-ControlNet, with DiT model weights initialized from FLUX.1-dev\cite{flux2024}. The model was trained on the GlyphMM-3M and Anyword-3M\cite{tuo2023anytext} datasets for 300,000 steps, and subsequently fine-tuned on the Poster-100K dataset for an additional 160,000 steps using 8 NVIDIA A800 GPUs. We employed a progressive training strategy: the glyph region loss was deactivated for the initial 100,000 steps, and a weighting parameter of $\lambda$ = 1 was applied after 100,000 steps. Since Flux.1-dev\cite{flux2024} supports training with images of arbitrary resolutions, we utilized the training images at their original resolutions. Optimization was performed using the AdamW\cite{adamw2019} optimizer with hyperparameters $\beta_1$ = 0.9 and $\beta_2$ = 0.999. We set an effective batch size of 16, a learning rate of 1e-5. Therefore, our effective training data volume amounts to 16$\times$ 460,000 = 7.36 million samples, which is substantially lower than the training data requirements of other visual text generation models, such as AnyText, which trained on 3M data for 10 epochs, and GlyphControl and TextDiffuser, which were trained on datasets exceeding tens of millions of samples. This outcome implicitly demonstrates the efficiency of our training strategy and methodology, as achieving competitive performance with limited computational resources and reduced data consumption underscores the effectiveness of our approach. Due to resource constraints, we only utilized a subset of our constructed dataset for training. This further validates that our strategy achieves high sample efficiency while maintaining robustness in text rendering tasks.

\subsection{Quantitative Results}
We comprehensively evaluate our model against specialized visual glyph generation approaches—including GlyphControl\cite{Yang_GlyphControl_Corr23}, AnyText\cite{tuo2023anytext}, AnyText2\cite{tuo2024anytext2}, GlyphDraw2\cite{glyphdraw2}, and Glyph-ByT5-v2\cite{glyphbyt5-v2}—along with baseline models SD3\cite{Patrick_SD3_ICML24} and ControlNet\cite{Zhang_ControlNet_Corr23}. To ensure comparability, all models used 40 sampling steps, with ControlNet-based architectures employing a control scale of 0.9. We fixed the sample count at 4, CFG scale at 0.9 where applicable, and sampling seed at 0. For GlyphDraw2\cite{glyphdraw2}, we used our self-trained implementation due to unavailable official weights, while AnyText2\cite{tuo2024anytext2} results were directly sourced from its paper.
Table \ref{tab:resultsontwobenchmark} validates UniGlyph's cross-lingual superiority across Chinese and English benchmarks. Our method achieves state-of-the-art performance in visual text generation while maintaining competitive text-semantic alignment.
Table \ref{tab:resultsonminitext} demonstrates UniGlyph's breakthrough in small-area text generation. While baseline models fail fundamentally and existing solutions show partial success with compromised textual integrity, our approach resolves the accuracy-coherence trade-off through joint optimization of semantic accuracy, structural preservation, and visual coherence. Remarkably, this advancement is achieved despite our limited training resources, further emphasizing the method's parameter efficiency and architectural superiority.
\subsection{Qualitative Results}
\begin{figure*}
    \centering
    \includegraphics[width=1\linewidth]{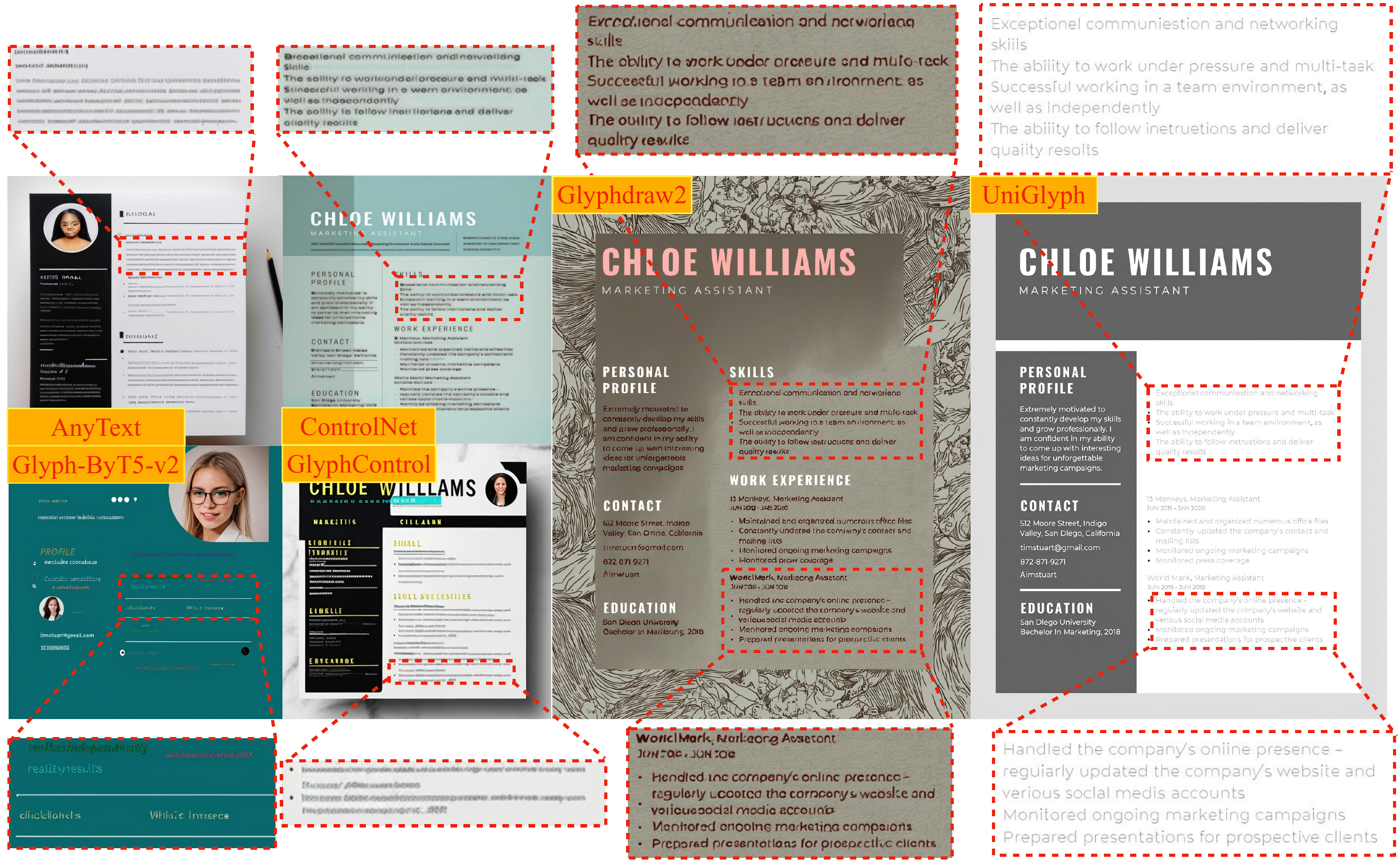}
    \caption{Comparative results of UniGlyph and other state-of-the-art models in small-character generation, highlighting UniGlyph’s exceptional performance in generating text within small regions. }
    \label{fig:qualires-smallglyph}
\end{figure*}

In our qualitative analysis, we selected several English test cases from the GlyphMM-benchmark. Previously, in our quantitative analysis, we had tested the model’s ability to generate small Chinese text using the MiniText-benchmark. Therefore, for the qualitative analysis, we chose to focus more on English test cases. The English data in our test set are highly complex, featuring many instances of small text regions and texts with intricate layouts. Also, because English letters are spaced more closely together than Chinese characters, generating small English text is a greater challenge for the model.

From the results in the first row, we observe that all models can correctly generate larger English letters in the center of the image. However, for the small text at the bottom of the image, only UniGlyph successfully and accurately renders all the letters. GlyphDraw2 also performs well, but when generating the small letters in the last row, it shows issues where letters blend together. This is the main reason for its lower accuracy in generating small characters.

As shown in the other examples, UniGlyph can generate curved text and produce multilingual text at the same time while maintaining the accuracy of small text regions. Additionally, the backgrounds it generates are visually appealing, which enhances the overall quality of the images.

To push the model to its limits in generating small glyphs, we selected a test case that contains the highest number of small glyph regions, as shown in Figure \ref{fig:qualires-smallglyph}. Both quantitative and qualitative evaluations demonstrate that our model has exceptional capability in generating small glyphs in both Chinese and English.

\subsection{Ablation Study}
In this section, we examine the impact of the adaptive glyph instruction strategy and the weighting factor $\lambda$ of the glyph region loss on our model’s performance. Given that the images in our GlyphMM-3M dataset are all high-resolution, we opted to conduct our ablation studies of $\lambda$ on the Anyword\cite{tuo2023anytext} dataset at a reduced resolution of 512$\times$512 to alleviate computational costs. To evaluate the extent of improvement on small glyph regions after applying the adaptive glyph condition, we continued to use the GlyphMM-3M training set, as it contains a greater number of small text regions. For each experimental setting, we consistently trained the model for 160,000 steps with an effective batch size of 2, which required approximately 40 hours on eight NVIDIA A800 GPUs. The training configurations remained consistent with those described in Sec.~\ref{sec:experiment}.

\begin{table}
\centering
\caption{Ablation experiments of different $\lambda$ values.}
\label{tab:alation-lambda}
\begin{tabular}{cccc}
\toprule
\textbf{$\lambda=$} & \textbf{Sen.Acc} & \textbf{NED} & \textbf{ClipScore} \\
\midrule
0       & 0.8179 & 0.8952 & 0.7868 \\
0.01    & 0.8161 & 0.8948 & 0.7876 \\
0.1     & 0.8166 & 0.8945 & 0.7871 \\
1       & \textbf{0.8188} & \textbf{0.8958} & \textbf{0.7896} \\
2       & 0.8182 & 0.8954 & 0.7876 \\
4       & 0.8158 & 0.8949 & 0.7870 \\
\bottomrule
\end{tabular}
\end{table}

\noindent We conducted six experiments with different values of $\lambda$, based on the results of our ablation experiments in Table \ref{tab:alation-lambda}, wWe validated the effectiveness of the glyph region loss by comparing scenarios where $\lambda$ equals 0 with those where $\lambda$ is non-zero and we determined that $\lambda$=1 yields the best results. Lastly, from the observed changes in the ClipScore, we found that slightly increasing the weight of $\lambda$ does not adversely affect the quality of image generation. Therefore, we do not require additional methods to enhance the generation of the background. In practical applications, the glyph region loss requires reconstructing images from the latent space to the pixel space, which can slow down training. Therefore, we apply the glyph region loss only after 100,000 training steps.

\begin{table}
\centering
\caption{Ablation experiments validated the necessity of constructing Adaptive Glyph Conditions (AGC) and evaluating the effectiveness of the strategies employed—specifically, the adaptive blending strategy and Gaussian blur}
\label{tab:ablation-adaptive}
\begin{tabular}{cccc}
\toprule 
\textbf{Method} & \textbf{Sen.Acc} & \textbf{NED} & \textbf{ClipScore} \\
\midrule
w/o AGC  & 0.7724 & 0.9348 & 0.8064 \\
w/o Gaussian Blur & 0.7851 & 0.9508 & 0.7963 \\
UniGlyph & \textbf{0.7849} & \textbf{0.9507} & \textbf{0.8097} \\
\bottomrule
\end{tabular}
\end{table}

\noindent To validate the necessity of constructing the Adaptive Glyph Condition (AGC), we conducted the experiments presented in Table \ref{tab:ablation-adaptive}. Our primary objective is to address the issue of inaccurate segmentation in small-character regions. To investigate the extent of improvement our strategy brings to datasets containing small characters, we employed the MiniText-benchmark as the test set. Initially, we completely omitted the AGC, utilizing only the raw segmentation maps as conditions. Subsequently, we applied the adaptive blending strategy without Gaussian blur.

\noindent The experimental results demonstrate that, after implementing the adaptive blending strategy, there was an improvement in the generation accuracy of small characters, albeit with a decrease in the CLIP score. However, upon incorporating Gaussian blur, the CLIP score increased while the generation accuracy did not decline. These findings confirm the effectiveness and necessity of both strategies.
\section{Conclusion}
\label{sec:conclu}

The generation of precise visual text in diffusion models has long posed a fundamental challenge due to the irreversible information degradation in conventional glyph conditioning paradigms. While existing ControlNet-based methods achieve initial success through multi-branch architectures and auxiliary modules, they inherently suffer from incomplete typographic signal preservation and architectural complexity that compromise text generation quality—particularly for small or stylized glyphs. Our proposed UniGlyph framework resolves this dilemma through a paradigm-shifting solution: replacing rendered glyph images with unified segmentation masks that preserve full typographic attributes at pixel resolution. This breakthrough eliminates the need for cascaded positional encoding and style control modules, enabling a single-ControlNet architecture to achieve state-of-the-art performance.

\noindent Quantitative evaluations across three benchmarks validate the framework's superiority. Crucially, our method achieves these advances without requiring specialized design for typographic attributes—a capability directly enabled by the complete signal preservation in segmentation masks.

\noindent Beyond architectural innovation, we propose the datasets GlyphMM-3M and Poster-100K, which alleviate the scarcity of high-resolution and high-quality training data in the existing visual text generation community. Additionally, our proposed test sets GlyphMM-benchmark and MiniText-benchmark facilitate a more comprehensive evaluation of the performance of visual text generation models.

\noindent Our results collectively demonstrate that preserving complete glyph signals through unified segmentation conditioning enables simultaneous architectural simplification and performance enhancement—a principle that redefines the design paradigm for visual text image generation. 
{
    \small
    \bibliographystyle{ieeenat_fullname}
    \bibliography{main}

\begin{thebibliography}{55}
\providecommand{\natexlab}[1]{#1}
\providecommand{\url}[1]{\texttt{#1}}
\expandafter\ifx\csname urlstyle\endcsname\relax
  \providecommand{\doi}[1]{doi: #1}\else
  \providecommand{\doi}{doi: \begingroup \urlstyle{rm}\Url}\fi

\bibitem[BlackForestLab(2024)]{FLUX.1}
BlackForestLab.
\newblock Flux.1.
\newblock \url{https://blackforestlabs.ai/announcing-black-forest-labs/}, 2024.

\bibitem[Chen et~al.(2023{\natexlab{a}})Chen, Huang, Lv, Cui, Chen, and Wei]{Chen_TextDiffuser2}
Jingye Chen, Yupan Huang, Tengchao Lv, Lei Cui, Qifeng Chen, and Furu Wei.
\newblock Textdiffuser-2: Unleashing the power of language models for text rendering.
\newblock \emph{arXiv preprint arXiv:2311.16465}, 2023{\natexlab{a}}.

\bibitem[Chen et~al.(2023{\natexlab{b}})Chen, Huang, Lv, Cui, Chen, and Wei]{Chen_TextDiffuser_Corr23}
Jingye Chen, Yupan Huang, Tengchao Lv, Lei Cui, Qifeng Chen, and Furu Wei.
\newblock Textdiffuser: Diffusion models as text painters.
\newblock \emph{arXiv preprint}, abs/2305.10855, 2023{\natexlab{b}}.

\bibitem[Chen et~al.(2023{\natexlab{c}})Chen, Pan, Li, Yao, and Mei]{chen2023control3d}
Yang Chen, Yingwei Pan, Yehao Li, Ting Yao, and Tao Mei.
\newblock Control3d: Towards controllable text-to-3d generation, 2023{\natexlab{c}}.

\bibitem[Cheng et~al.(2025)Cheng, Xiong, Wei, Zhu, and Yuan]{cheng2025whoever}
Runxi Cheng, Feng Xiong, Yongxian Wei, Wanyun Zhu, and Chun Yuan.
\newblock Whoever started the interference should end it: Guiding data-free model merging via task vectors.
\newblock \emph{arXiv preprint arXiv:2503.08099}, 2025.

\bibitem[Du et~al.(2024)Du, Zhang, Wang, Wang, Yue, Zhang, Ding, Wang, Xu, and Yuan]{du2024alore}
Sinan Du, Guosheng Zhang, Keyao Wang, Yuanrui Wang, Haixiao Yue, Gang Zhang, Errui Ding, Jingdong Wang, Zhengzhuo Xu, and Chun Yuan.
\newblock Alore: Efficient visual adaptation via aggregating low rank experts.
\newblock \emph{arXiv preprint arXiv:2412.08341}, 2024.

\bibitem[Esser et~al.(2024)Esser, Kulal, Blattmann, Entezari, M{\"{u}}ller, Saini, Levi, Lorenz, Sauer, Boesel, Podell, Dockhorn, English, and Rombach]{Patrick_SD3_ICML24}
Patrick Esser, Sumith Kulal, Andreas Blattmann, Rahim Entezari, Jonas M{\"{u}}ller, Harry Saini, Yam Levi, Dominik Lorenz, Axel Sauer, Frederic Boesel, Dustin Podell, Tim Dockhorn, Zion English, and Robin Rombach.
\newblock Scaling rectified flow transformers for high-resolution image synthesis.
\newblock In \emph{ICML}, 2024.

\bibitem[Feng et~al.(2023)Feng, Zhu, jui Fu, Jampani, Akula, He, Basu, Wang, and Wang]{feng2023layoutgpt}
Weixi Feng, Wanrong Zhu, Tsu jui Fu, Varun Jampani, Arjun Akula, Xuehai He, Sugato Basu, Xin~Eric Wang, and William~Yang Wang.
\newblock Layoutgpt: Compositional visual planning and generation with large language models, 2023.

\bibitem[Gu et~al.(2022)Gu, Meng, Lu, Hou, Niu, Xu, Liang, Zhang, Jiang, and Xu]{wukong_corr22}
Jiaxi Gu, Xiaojun Meng, Guansong Lu, Lu Hou, Minzhe Niu, Hang Xu, Xiaodan Liang, Wei Zhang, Xin Jiang, and Chunjing Xu.
\newblock Wukong: 100 million large-scale chinese cross-modal pre-training dataset and {A} foundation framework.
\newblock \emph{CoRR}, abs/2202.06767, 2022.

\bibitem[Guan et~al.(2025)Guan, Cheng, Liu, and Yuan]{guan2025enhancinglogitsdistillationplugplay}
Yuchen Guan, Runxi Cheng, Kang Liu, and Chun Yuan.
\newblock Enhancing logits distillation with plug\&play kendall's $\tau$ ranking loss, 2025.

\bibitem[Hu et~al.(2023)Hu, Gao, Zhang, Sun, Zhang, and Bo]{hu_animateanyone}
Li Hu, Xin Gao, Peng Zhang, Ke Sun, Bang Zhang, and Liefeng Bo.
\newblock Animate anyone: Consistent and controllable image-to-video synthesis for character animation.
\newblock \emph{CoRR}, 2023.

\bibitem[Huang et~al.(2023)Huang, Chen, Liu, Shen, Zhao, and Zhou]{Huang_Composer_Corr23}
Lianghua Huang, Di Chen, Yu Liu, Yujun Shen, Deli Zhao, and Jingren Zhou.
\newblock Composer: Creative and controllable image synthesis with composable conditions.
\newblock \emph{arXiv preprint}, abs/2302.09778, 2023.

\bibitem[Ji et~al.(2023)Ji, Zhang, Wang, Hou, Zhang, Price, and Chang]{diffste}
Jiabao Ji, Guanhua Zhang, Zhaowen Wang, Bairu Hou, Zhifei Zhang, Brian Price, and Shiyu Chang.
\newblock Improving diffusion models for scene text editing with dual encoders, 2023.

\bibitem[Kingma and Welling(2014)]{Diederik_VAE_ICLR14}
Diederik~P. Kingma and Max Welling.
\newblock Auto-encoding variational bayes.
\newblock In \emph{ICLR}, 2014.

\bibitem[Kirillov et~al.(2023)Kirillov, Mintun, Ravi, Mao, Rolland, Gustafson, Xiao, Whitehead, Berg, Lo, Doll{\'a}r, and Girshick]{kirillov2023segany}
Alexander Kirillov, Eric Mintun, Nikhila Ravi, Hanzi Mao, Chloe Rolland, Laura Gustafson, Tete Xiao, Spencer Whitehead, Alexander~C. Berg, Wan-Yen Lo, Piotr Doll{\'a}r, and Ross Girshick.
\newblock Segment anything.
\newblock \emph{arXiv:2304.02643}, 2023.

\bibitem[Labs(2024)]{flux2024}
Black~Forest Labs.
\newblock Flux.
\newblock \url{https://github.com/black-forest-labs/flux}, 2024.

\bibitem[Lin et~al.(2023)Lin, Guo, Sun, Yang, Lou, and Zhang]{lin2023layoutprompter}
Jiawei Lin, Jiaqi Guo, Shizhao Sun, Zijiang~James Yang, Jian-Guang Lou, and Dongmei Zhang.
\newblock Layoutprompter: Awaken the design ability of large language models, 2023.

\bibitem[Lipman et~al.(2023)Lipman, Chen, Ben-Hamu, Nickel, and Le]{lipman2023flowmatching}
Yaron Lipman, Ricky T.~Q. Chen, Heli Ben-Hamu, Maximilian Nickel, and Matt Le.
\newblock Flow matching for generative modeling, 2023.

\bibitem[Liu et~al.(2023)Liu, Garrette, Saharia, Chan, Roberts, Narang, Blok, Mical, Norouzi, and Constant]{Liu_CharacterAware_ACL23}
Rosanne Liu, Dan Garrette, Chitwan Saharia, William Chan, Adam Roberts, Sharan Narang, Irina Blok, RJ Mical, Mohammad Norouzi, and Noah Constant.
\newblock Character-aware models improve visual text rendering.
\newblock In \emph{ACL}, pages 16270--16297, 2023.

\bibitem[Liu et~al.(2024{\natexlab{a}})Liu, Liang, Liang, Luo, Li, Huang, and Yuan]{glyphbyt5}
Zeyu Liu, Weicong Liang, Zhanhao Liang, Chong Luo, Ji Li, Gao Huang, and Yuhui Yuan.
\newblock Glyph-byt5: A customized text encoder for accurate visual text rendering.
\newblock \emph{arXiv preprint arXiv:2403.09622}, 2024{\natexlab{a}}.

\bibitem[Liu et~al.(2024{\natexlab{b}})Liu, Liang, Zhao, Chen, Li, and Yuan]{glyphbyt5-v2}
Zeyu Liu, Weicong Liang, Yiming Zhao, Bohan Chen, Ji Li, and Yuhui Yuan.
\newblock Glyph-byt5-v2: A strong aesthetic baseline for accurate multilingual visual text rendering.
\newblock \emph{arXiv preprint arXiv:2406.10208}, 2024{\natexlab{b}}.

\bibitem[Loshchilov and Hutter(2019)]{adamw2019}
Ilya Loshchilov and Frank Hutter.
\newblock Decoupled weight decay regularization, 2019.

\bibitem[Ma et~al.(2024{\natexlab{a}})Ma, Deng, Chen, Lu, and Yang]{glyphdraw2}
Jian Ma, Yonglin Deng, Chen Chen, Haonan Lu, and Zhenyu Yang.
\newblock Glyphdraw2: Automatic generation of complex glyph posters with diffusion models and large language models.
\newblock \emph{CoRR}, 2024{\natexlab{a}}.

\bibitem[Ma et~al.(2024{\natexlab{b}})Ma, Yue, Fu, Zhong, Zhou, Wei, and Hu]{ma2024chargenhighaccuratecharacterlevel}
Lichen Ma, Tiezhu Yue, Pei Fu, Yujie Zhong, Kai Zhou, Xiaoming Wei, and Jie Hu.
\newblock Chargen: High accurate character-level visual text generation model with multimodal encoder, 2024{\natexlab{b}}.

\bibitem[Ma et~al.(2024{\natexlab{c}})Ma, Liu, Wang, Pan, He, Yuan, Zeng, Cai, Shum, Liu, et~al.]{ma2024followyouremoji}
Yue Ma, Hongyu Liu, Hongfa Wang, Heng Pan, Yingqing He, Junkun Yuan, Ailing Zeng, Chengfei Cai, Heung-Yeung Shum, Wei Liu, et~al.
\newblock Follow-your-emoji: Fine-controllable and expressive freestyle portrait animation.
\newblock In \emph{SIGGRAPH Asia 2024 Conference Papers}, pages 1--12, 2024{\natexlab{c}}.

\bibitem[Ma et~al.(2025{\natexlab{a}})Ma, Feng, Zhang, Liu, Zhang, Xing, Zhang, Yang, Wang, and Chen]{ma2025followcreation}
Yue Ma, Kunyu Feng, Xinhua Zhang, Hongyu Liu, David~Junhao Zhang, Jinbo Xing, Yinhan Zhang, Ayden Yang, Zeyu Wang, and Qifeng Chen.
\newblock Follow-your-creation: Empowering 4d creation through video inpainting.
\newblock \emph{arXiv preprint arXiv:2506.04590}, 2025{\natexlab{a}}.

\bibitem[Ma et~al.(2025{\natexlab{b}})Ma, He, Wang, Wang, Shen, Qi, Ying, Cai, Li, Shum, et~al.]{ma2025followyourclick}
Yue Ma, Yingqing He, Hongfa Wang, Andong Wang, Leqi Shen, Chenyang Qi, Jixuan Ying, Chengfei Cai, Zhifeng Li, Heung-Yeung Shum, et~al.
\newblock Follow-your-click: Open-domain regional image animation via motion prompts.
\newblock In \emph{Proceedings of the AAAI Conference on Artificial Intelligence}, pages 6018--6026, 2025{\natexlab{b}}.

\bibitem[Ma et~al.(2025{\natexlab{c}})Ma, Liu, Zhu, Yang, Feng, Zhang, Li, Han, Qi, and Chen]{ma2025followyourmotion}
Yue Ma, Yulong Liu, Qiyuan Zhu, Ayden Yang, Kunyu Feng, Xinhua Zhang, Zhifeng Li, Sirui Han, Chenyang Qi, and Qifeng Chen.
\newblock Follow-your-motion: Video motion transfer via efficient spatial-temporal decoupled finetuning.
\newblock \emph{arXiv preprint arXiv:2506.05207}, 2025{\natexlab{c}}.

\bibitem[Mou et~al.(2023)Mou, Wang, Xie, Zhang, Qi, Shan, and Qie]{Mou_T2Iadapter_Corr23}
Chong Mou, Xintao Wang, Liangbin Xie, Jian Zhang, Zhongang Qi, Ying Shan, and Xiaohu Qie.
\newblock T2i-adapter: Learning adapters to dig out more controllable ability for text-to-image diffusion models.
\newblock \emph{arXiv preprint}, abs/2302.08453, 2023.

\bibitem[OpenAI(2023)]{DALLE3}
OpenAI.
\newblock Dall·e3.
\newblock \url{https://openai.com/index/dall-e-3/}, 2023.

\bibitem[PaddlePaddle(2023)]{PP-OCRv4}
PaddlePaddle.
\newblock Pp-ocrv4.
\newblock \url{https://github.com/PaddlePaddle/PaddleOCR/blob/release/2.7/doc/doc_ch/PP-OCRv4_introduction.md}, 2023.

\bibitem[Paliwal et~al.(2024)Paliwal, Jain, Sharma, Jamwal, and Vig]{paliwal2024customtext}
Shubham Paliwal, Arushi Jain, Monika Sharma, Vikram Jamwal, and Lovekesh Vig.
\newblock Customtext: Customized textual image generation using diffusion models, 2024.

\bibitem[Peebles and Xie(2023)]{peebles2023dit}
William Peebles and Saining Xie.
\newblock Scalable diffusion models with transformers, 2023.

\bibitem[Schuhmann et~al.(2021)Schuhmann, Vencu, Beaumont, Kaczmarczyk, Mullis, Katta, Coombes, Jitsev, and Komatsuzaki]{laion_400m_corr21}
Christoph Schuhmann, Richard Vencu, Romain Beaumont, Robert Kaczmarczyk, Clayton Mullis, Aarush Katta, Theo Coombes, Jenia Jitsev, and Aran Komatsuzaki.
\newblock {LAION-400M:} open dataset of clip-filtered 400 million image-text pairs.
\newblock \emph{CoRR}, abs/2111.02114, 2021.

\bibitem[Sun et~al.(2024)Sun, Shu, Zhou, Yu, Chen, Yang, and Chun]{sun2024gsrenderdeduplicatedoccupancyprediction}
Qianpu Sun, Changyong Shu, Sifan Zhou, Zichen Yu, Yan Chen, Dawei Yang, and Yuan Chun.
\newblock Gsrender: Deduplicated occupancy prediction via weakly supervised 3d gaussian splatting, 2024.

\bibitem[Touvron et~al.(2023{\natexlab{a}})Touvron, Lavril, Izacard, Martinet, Lachaux, Lacroix, Rozière, Goyal, Hambro, Azhar, Rodriguez, Joulin, Grave, and Lample]{touvron2023llama}
Hugo Touvron, Thibaut Lavril, Gautier Izacard, Xavier Martinet, Marie-Anne Lachaux, Timothée Lacroix, Baptiste Rozière, Naman Goyal, Eric Hambro, Faisal Azhar, Aurelien Rodriguez, Armand Joulin, Edouard Grave, and Guillaume Lample.
\newblock Llama: Open and efficient foundation language models, 2023{\natexlab{a}}.

\bibitem[Touvron et~al.(2023{\natexlab{b}})Touvron, Martin, Stone, Albert, Almahairi, Babaei, Bashlykov, Batra, Bhargava, Bhosale, Bikel, Blecher, Ferrer, Chen, Cucurull, Esiobu, Fernandes, Fu, Fu, Fuller, Gao, Goswami, Goyal, Hartshorn, Hosseini, Hou, Inan, Kardas, Kerkez, Khabsa, Kloumann, Korenev, Koura, Lachaux, Lavril, Lee, Liskovich, Lu, Mao, Martinet, Mihaylov, Mishra, Molybog, Nie, Poulton, Reizenstein, Rungta, Saladi, Schelten, Silva, Smith, Subramanian, Tan, Tang, Taylor, Williams, Kuan, Xu, Yan, Zarov, Zhang, Fan, Kambadur, Narang, Rodriguez, Stojnic, Edunov, and Scialom]{touvron2023llama2}
Hugo Touvron, Louis Martin, Kevin Stone, Peter Albert, Amjad Almahairi, Yasmine Babaei, Nikolay Bashlykov, Soumya Batra, Prajjwal Bhargava, Shruti Bhosale, Dan Bikel, Lukas Blecher, Cristian~Canton Ferrer, Moya Chen, Guillem Cucurull, David Esiobu, Jude Fernandes, Jeremy Fu, Wenyin Fu, Brian Fuller, Cynthia Gao, Vedanuj Goswami, Naman Goyal, Anthony Hartshorn, Saghar Hosseini, Rui Hou, Hakan Inan, Marcin Kardas, Viktor Kerkez, Madian Khabsa, Isabel Kloumann, Artem Korenev, Punit~Singh Koura, Marie-Anne Lachaux, Thibaut Lavril, Jenya Lee, Diana Liskovich, Yinghai Lu, Yuning Mao, Xavier Martinet, Todor Mihaylov, Pushkar Mishra, Igor Molybog, Yixin Nie, Andrew Poulton, Jeremy Reizenstein, Rashi Rungta, Kalyan Saladi, Alan Schelten, Ruan Silva, Eric~Michael Smith, Ranjan Subramanian, Xiaoqing~Ellen Tan, Binh Tang, Ross Taylor, Adina Williams, Jian~Xiang Kuan, Puxin Xu, Zheng Yan, Iliyan Zarov, Yuchen Zhang, Angela Fan, Melanie Kambadur, Sharan Narang, Aurelien Rodriguez, Robert Stojnic, Sergey Edunov, and Thomas
  Scialom.
\newblock Llama 2: Open foundation and fine-tuned chat models, 2023{\natexlab{b}}.

\bibitem[Tuo et~al.(2023)Tuo, Xiang, He, Geng, and Xie]{tuo2023anytext}
Yuxiang Tuo, Wangmeng Xiang, Jun-Yan He, Yifeng Geng, and Xuansong Xie.
\newblock Anytext: Multilingual visual text generation and editing.
\newblock \emph{arXiv}, 2023.

\bibitem[Tuo et~al.(2024)Tuo, Geng, and Bo]{tuo2024anytext2}
Yuxiang Tuo, Yifeng Geng, and Liefeng Bo.
\newblock Anytext2: Visual text generation and editing with customizable attributes, 2024.

\bibitem[Wang et~al.(2024)Wang, Yang, Huang, Wang, and Li]{wang2024rectifieddiffusion}
Fu-Yun Wang, Ling Yang, Zhaoyang Huang, Mengdi Wang, and Hongsheng Li.
\newblock Rectified diffusion: Straightness is not your need in rectified flow, 2024.

\bibitem[Wei et~al.(2025{\natexlab{a}})Wei, Liu, Yuan, and Zhang]{wei2025perceiveunderstandrestorerealworld}
Hongyang Wei, Shuaizheng Liu, Chun Yuan, and Lei Zhang.
\newblock Perceive, understand and restore: Real-world image super-resolution with autoregressive multimodal generative models, 2025{\natexlab{a}}.

\bibitem[Wei et~al.(2025{\natexlab{b}})Wei, Tang, Shen, Hu, Yuan, and Cao]{wei2025modeling}
Yongxian Wei, Anke Tang, Li Shen, Zixuan Hu, Chun Yuan, and Xiaochun Cao.
\newblock Modeling multi-task model merging as adaptive projective gradient descent.
\newblock \emph{arXiv preprint arXiv:2501.01230}, 2025{\natexlab{b}}.

\bibitem[Xu et~al.(2022)Xu, Qi, Ma, Zhang, Shan, and Qie]{xu2022bts}
Xixi Xu, Zhongang Qi, Jianqi Ma, Honglun Zhang, Ying Shan, and Xiaohu Qie.
\newblock Bts: a bi-lingual benchmark for text segmentation in the wild.
\newblock In \emph{Proceedings of the IEEE/CVF conference on computer vision and pattern recognition}, pages 19152--19162, 2022.

\bibitem[Xu et~al.(2023)Xu, Du, Qi, Xu, Yuan, and Guo]{xu2023chartbench}
Zhengzhuo Xu, Sinan Du, Yiyan Qi, Chengjin Xu, Chun Yuan, and Jian Guo.
\newblock Chartbench: A benchmark for complex visual reasoning in charts.
\newblock \emph{arXiv preprint arXiv:2312.15915}, 2023.

\bibitem[Xu et~al.(2025)Xu, Qu, Qi, Du, Xu, Yuan, and Guo]{xu2025chartmoe}
Zhengzhuo Xu, Bowen Qu, Yiyan Qi, SiNan Du, Chengjin Xu, Chun Yuan, and Jian Guo.
\newblock Chartmoe: Mixture of diversely aligned expert connector for chart understanding.
\newblock In \emph{The Thirteenth International Conference on Learning Representations}, 2025.

\bibitem[Yang et~al.(2023)Yang, Gui, Yuan, Ding, Hu, and Chen]{Yang_GlyphControl_Corr23}
Yukang Yang, Dongnan Gui, Yuhui Yuan, Haisong Ding, Han Hu, and Kai Chen.
\newblock Glyphcontrol: Glyph conditional control for visual text generation.
\newblock \emph{arXiv preprint}, abs/2305.18259, 2023.

\bibitem[Ye et~al.(2023)Ye, Zhang, Liu, Han, and Yang]{ip-adapter}
Hu Ye, Jun Zhang, Sibo Liu, Xiao Han, and Wei Yang.
\newblock Ip-adapter: Text compatible image prompt adapter for text-to-image diffusion models.
\newblock \emph{arXiv preprint}, 2023.

\bibitem[Ye et~al.(2024)Ye, Zhang, Liu, Liu, Yin, Liu, Du, and Tao]{hi-sam_tpami24}
Maoyuan Ye, Jing Zhang, Juhua Liu, Chenyu Liu, Baocai Yin, Cong Liu, Bo Du, and Dacheng Tao.
\newblock Hi-sam: Marrying segment anything model for hierarchical text segmentation.
\newblock \emph{IEEE Transactions on Pattern Analysis and Machine Intelligence}, pages 1--16, 2024.

\bibitem[Zhang et~al.(2024)Zhang, Zhou, Gu, Wigington, Yu, Chen, Sun, and Zhang]{zhang2024artist}
Jianyi Zhang, Yufan Zhou, Jiuxiang Gu, Curtis Wigington, Tong Yu, Yiran Chen, Tong Sun, and Ruiyi Zhang.
\newblock Artist: Improving the generation of text-rich images with disentangled diffusion models and large language models, 2024.

\bibitem[Zhang and Agrawala(2023)]{Zhang_ControlNet_Corr23}
Lvmin Zhang and Maneesh Agrawala.
\newblock Adding conditional control to text-to-image diffusion models.
\newblock \emph{arXiv preprint}, abs/2302.05543, 2023.

\bibitem[Zhang et~al.(2023)Zhang, Chen, Wang, Lu, and Qiao]{zhang2023_brushyourtext}
Lingjun Zhang, Xinyuan Chen, Yaohui Wang, Yue Lu, and Yu Qiao.
\newblock Brush your text: Synthesize any scene text on images via diffusion model, 2023.

\bibitem[Zhangli et~al.(2024)Zhangli, Jiang, Liu, Yu, Dai, Ramchandani, Pang, Metaxas, and Krishnan]{zhangli2024layoutagnosticscenetext}
Qilong Zhangli, Jindong Jiang, Di Liu, Licheng Yu, Xiaoliang Dai, Ankit Ramchandani, Guan Pang, Dimitris~N. Metaxas, and Praveen Krishnan.
\newblock Layout agnostic scene text image synthesis with diffusion models, 2024.

\bibitem[Zhao and Lian(2023)]{zhao_udifftex}
Yiming Zhao and Zhouhui Lian.
\newblock Udifftext: A unified framework for high-quality text synthesis in arbitrary images via character-aware diffusion models, 2023.

\bibitem[Zhuang et~al.(2023)Zhuang, Zeng, Liu, Yuan, and Chen]{zhuang2023task}
Junhao Zhuang, Yanhong Zeng, Wenran Liu, Chun Yuan, and Kai Chen.
\newblock A task is worth one word: Learning with task prompts for high-quality versatile image inpainting, 2023.

\bibitem[Zhuang et~al.(2024)Zhuang, Ju, Zhang, Liu, Zhang, Yuan, and Shan]{zhuang2024colorflow}
Junhao Zhuang, Xuan Ju, Zhaoyang Zhang, Yong Liu, Shiyi Zhang, Chun Yuan, and Ying Shan.
\newblock Colorflow: Retrieval-augmented image sequence colorization, 2024.

\end{thebibliography}
}


\end{document}